\documentclass[letterpaper]{article}

\IfFileExists{aaai2027.sty}{%
}{%
  \usepackage{aaai2027_fallback}%
}

\usepackage{times}
\usepackage{helvet}
\usepackage{courier}
\usepackage{algorithm}
\usepackage{algorithmic}
\usepackage[hyphens]{url}
\usepackage{graphicx}
\usepackage{natbib}
\usepackage{booktabs}
\usepackage{amsmath}
\usepackage{amssymb}
\usepackage{xspace}
\usepackage{bm}
\title{PEG-Tab: Sampling-Time Record Repair and Release Control for
Tabular Synthesis}

\author{
Pengfei Li\textsuperscript{\rm 1},
Qinyi Liu\textsuperscript{\rm 2},
Mohammad Khalil\textsuperscript{\rm 1}
}

\affiliations{
\textsuperscript{\rm 1}Centre for the Science of Learning \& Technology (SLATE), University of Bergen, Bergen, Norway\\
\textsuperscript{\rm 2}School of Education, City University of Macau, Macao SAR, China
}

\begin{document}
\maketitle

\begin{abstract}
Pretrained tabular generators can reproduce training records even when
aggregate utility remains high. When retraining is unavailable or too
costly, sampling and release are the remaining intervention points. We
present PEG-Tab (Post-Training Energy Guidance for Tabular Synthesis), a
post-training repair and release-control framework for frozen tabular
generators. For each generated row, a generator-native operator creates
two alternatives. A shared calibrated score compares the three
candidates, favours lower-risk records, and applies a final release
check. We instantiate this interface for GReaT, CTGAN, TVAE, and
TabDDPM without updating their parameters. Across five datasets and
four generator families, PEG-Tab reduces mean Near Copy from $0.078$ to
$0.027$ and lowers aggregate Exact Copy to zero. Relative to a
$3\times$ post hoc filter, it retains higher utility in 12 of 16
transfer settings and Pareto-dominates the filter in eight. Gains are
concentrated in copy and proximity-related risks. 
\end{abstract}

\section{Introduction}

Synthetic tables support data sharing, model development, and restricted analysis when access to original records is limited \cite{cormode2025survey}. Representative generators include Conditional Tabular GAN (CTGAN), Tabular Variational Autoencoder (TVAE), Generation of Realistic Tabular Data (GReaT), and Tabular Denoising Diffusion Probabilistic Model (TabDDPM) \cite{xu2019modeling,borisov2023language,kotelnikov2023tabddpm}. Strong population-level fidelity does not preclude record-level memorisation. A generator may reproduce a training row, emit a close neighbour, or place unusual mass around rare records \cite{carlini2019secret,hyeong2022empirical,vanbreugel2023membership}.

These risks motivate interventions before synthetic records are
released. Existing options operate at different stages. Post-generation
methods can filter or refine completed synthetic data, but they do not
modify the sampling trajectory that produced a risky record
\citep{wang2023post}. Training a generator with differential privacy
(DP) provides a formal guarantee, but requires control of the training
pipeline and introduces a nontrivial privacy--utility trade-off
\citep{ponomareva2023dpfy,chen2025benchmarking}. We instead consider a
setting in which an organisation already has a trained generator and
retains access to its inference process, but cannot or does not wish to
retrain it. The original infrastructure, optimiser state, or retraining
budget may no longer be available. This leads to our central question:
can risky rows be repaired before release while the generator remains
fixed?

\begin{figure*}[t]
    \centering
    \includegraphics[width=0.99\textwidth]{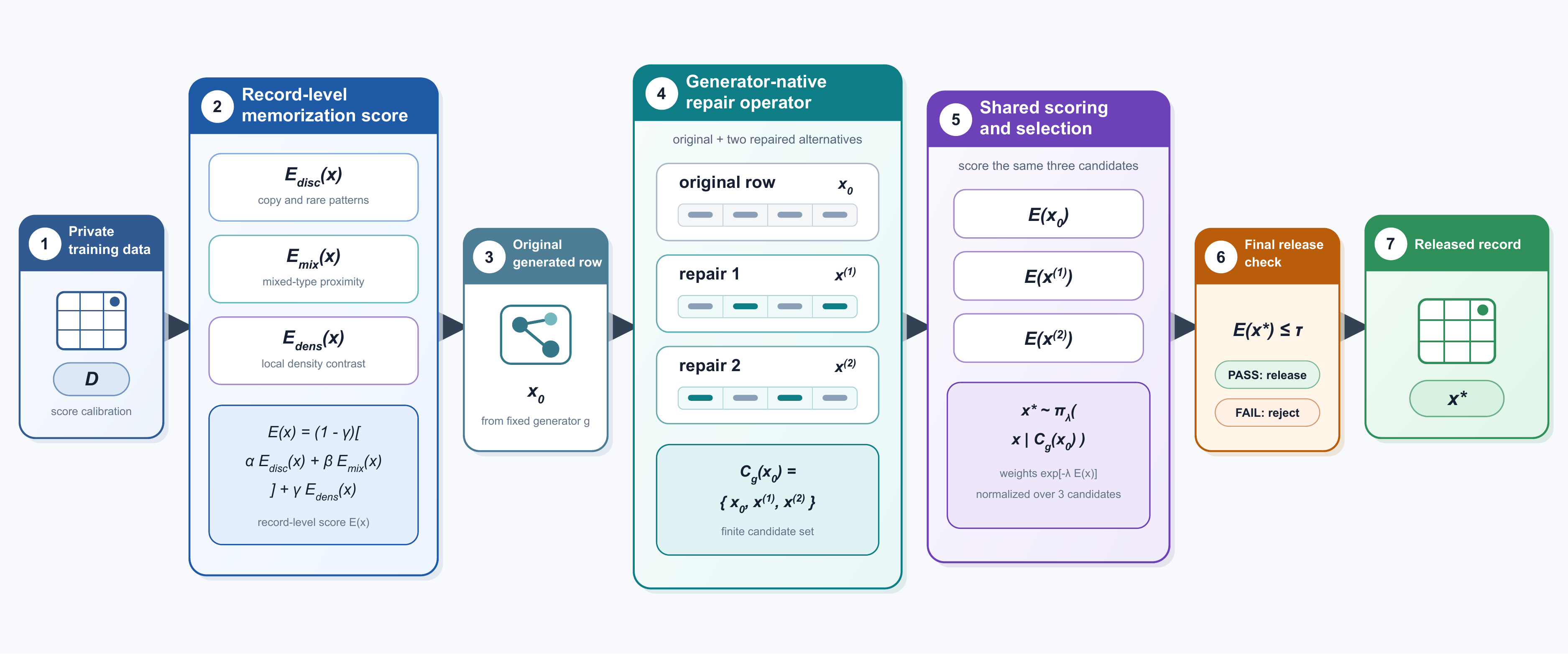}
    \caption{PEG-Tab repairs and checks a generated record before release. Offline statistics define the record-level risk score. Each generator then uses its native sampling mechanism to produce alternatives. The same scoring and release rule is applied to the original row and the repaired candidates. The base generator is not retrained, and the score is not a formal privacy guarantee.}
    \label{fig:overview}
\end{figure*}

We propose PEG-Tab (Post-training Energy Guidance for Tabular Synthesis), a sampling-time record repair and release-control framework. PEG-Tab decouples generator-native editing from a shared candidate score and release gate, as illustrated in Figure~\ref{fig:overview}. For each generated row, the native mechanism produces two alternatives. PEG-Tab scores the resulting three-record set, favours lower-score candidates, and verifies the selected row before release. GReaT uses risk-aware decoding. CTGAN and TVAE repair latent representations. TabDDPM uses masked reverse regeneration. The same control logic therefore spans language-model, adversarial, variational, and diffusion generators without weight updates.

The record score combines direct copying, rarity, mixed-space proximity, and local density. It is an operational decision score used to compare candidates under a single fixed calibration regime. It is not a new privacy definition, a membership probability, or a formal guarantee. We therefore separate direct copy outcomes and score-aligned diagnostics from a held-out Shadow membership inference attack (MIA) excluded from score construction and tuning.

Our contributions are threefold. First, we formulate post-training tabular release as finite-candidate repair, shared record scoring, and final gating, separating model-specific editing from model-independent release control. Second, we instantiate this interface for four heterogeneous generator families without parameter updates. Third, we evaluate a predefined development-to-transfer protocol with paired uncertainty, a held-out attack, pipeline ablations, post hoc filtering, formal DP reference mechanisms, parameter sensitivity, and sampling cost. Under frozen transfer settings, Near Copy never worsens across the 16 cells. Unchanged DOMIAS and Shadow MIA results bound the supported claim to copy-style memorisation and related proximity risks.

\section{Related Work}

\paragraph{Tabular synthesis.}
CTGAN and TVAE adapt adversarial and variational generation to mixed-type tables \cite{xu2019modeling}. GReaT serialises rows and fine-tunes an autoregressive language model \cite{borisov2023language}. TabDDPM models continuous and categorical columns through diffusion \cite{kotelnikov2023tabddpm}. Their sampling procedures differ enough that one model-internal update is difficult to reuse across all four families.

\paragraph{Memorization audits.}
Synthetic data can reveal training membership and record similarity even when marginal fidelity is high \cite{carlini2019secret,hyeong2022empirical}. DOMIAS detects local overfitting through density differences \cite{vanbreugel2023membership}. DPI studies copying in tabular generators \cite{ward2024dpi}. Similarity measures are useful diagnostics, but are not privacy guarantees and can miss other attacks \cite{ganev2025inadequacy}. We therefore distinguish score-aligned diagnostics from a held-out Shadow MIA.

\paragraph{Private synthesis and release-time control.}
DP-SGD protects training through clipping and noise \cite{abadi2016deep}, while MST and AIM generate tables from differentially private measurements \cite{mckenna2021mst,mckenna2022aim}. Recent work also studies private language-model synthesis and private population refinement \cite{tran2024dpllmtgen,tran2026tabpe}. These methods redesign training or population construction. PEG-Tab instead keeps an existing generator fixed and controls individual rows during sampling.

Post hoc filtering is the closest low-cost alternative. It discards high-risk rows from an oversized pool, while PEG-Tab tests whether native repair can recover a releasable row. Its operators build on controlled decoding, gradient guidance, and masked diffusion repair \cite{dathathri2020pplm,dhariwal2021diffusion,lugmayr2022repaint}. The contribution is a common record-level control interface across heterogeneous frozen generators.

\section{Method}

\subsection{Setting}
Let $\mathcal{D}=\{x_i\}_{i=1}^{n}$ be a private mixed-type table and let $G_\theta$ be a pretrained generator. The data holder runs PEG-Tab internally and releases only the resulting synthetic table. The method requires the private training table and access to the generator's inference process. This means token logits for GReaT, latent and decoder access for CTGAN and TVAE, and reverse-process access for TabDDPM. The model weights remain fixed.

The base generator first produces a row $x_0$. PEG-Tab checks that row, repairs the fields that contribute most to its risk, and chooses one record for release from the original and two repaired alternatives. Figure~\ref{fig:overview} summarises the workflow. PEG-Tab is operated internally by the data holder and requires more than black-box access. Our goal is to reduce direct copying and unusual concentration around training records. We do not claim differential privacy or protection against every attack class.

\subsection{Record-Level Risk Score}
PEG-Tab uses a score that summarises observable forms of record memorisation:
\begin{equation}
\label{eq:energy}
\begin{split}
E(x) &= (1-\gamma)\alpha E_{\mathrm{disc}}(x) + (1-\gamma)\beta E_{\mathrm{mix}}(x) \\
     &\quad + \gamma E_{\mathrm{dens}}(x).
\end{split}
\end{equation}
Each component is calibrated to $[0,1]$ with reference quantiles. The same calibrated transform is applied to every candidate within a run, and calibration does not use attack outcomes. The combined score is used only within that regime and need not itself lie in $[0,1]$. A population-level fidelity measure asks whether a synthetic table resembles the real distribution. In contrast, $E(x)$ asks whether one candidate row is unusually tied to particular training records. It is not an exposure estimate, a membership probability, or a formal privacy guarantee.

\paragraph{Discrete copying and rarity.}
For a serialised or categorical record, let $\mathcal{G}(x)$ denote value substrings and let $\mathcal{K}(x)$ denote selected column combinations. We use
\begin{equation}
\widetilde E_{\mathrm{disc}}(x)=
\mathbb{1}[x\in\mathcal{D}]+\frac{1}{Z_x}
\left(\sum_{g\in\mathcal{G}(x)}r_g+\sum_{k\in\mathcal{K}(x)}r_k\right),
\label{eq:disc}
\end{equation}
where $r_g$ and $r_k$ increase as the corresponding value or combination becomes rarer in $\mathcal{D}$. $Z_x$ normalises by the number of active terms. Calibration of $\widetilde E_{\mathrm{disc}}$ gives $E_{\mathrm{disc}}$.

\paragraph{Mixed-space proximity and rarity.}
Distances use a normalised mixed-type representation. Let $d_1(x,\mathcal{D})$ be the closest training distance. We combine bounded nearest-neighbour proximity with binned and cross-feature rarity:
\begin{equation}
\begin{split}
E_{\mathrm{mix}}(x)={}&\frac{\omega_{\mathrm{nn}}}{1+d_1(x,\mathcal{D})}
+\omega_{\mathrm{bin}}s_{\mathrm{bin}}(x)\\
&+\omega_{\mathrm{comb}}s_{\mathrm{comb}}(x).
\end{split}
\label{eq:mixed}
\end{equation}
The two rarity terms capture unusual continuous bins and mixed-feature combinations. A separate density contrast compares mean local distances to training and reference records:
\begin{equation}
E_{\mathrm{dens}}(x)=\max\left(0,1-
\frac{\bar d_{\mathrm{train}}(x)}{\bar d_{\mathrm{ref}}(x)+\epsilon}\right).
\label{eq:density}
\end{equation}
Reference records are excluded from generator training. When no suitable reference population exists, the density component can be disabled by setting $\gamma=0$.

\begin{table}[t]
\centering
\small
\setlength{\tabcolsep}{3.5pt}
\begin{tabular}{lp{2.0cm}p{3.0cm}}
\toprule
Term & Large when & Intended signal \\
\midrule
$E_{\mathrm{disc}}$
& Values or combinations are copied or rare
& Direct reproduction and uniqueness \\

$E_{\mathrm{mix}}$
& A candidate is close to a member or occupies a rare mixed-space region
& Neighbourhood overlap and mixed-feature rarity \\

$E_{\mathrm{dens}}$
& Training neighbours are closer than reference neighbours
& Local concentration around training data \\
\bottomrule
\end{tabular}

\caption{Meaning of the record-level score components. Each term describes one candidate row and does not constitute a population-level privacy guarantee.}
\label{tab:score_semantics}
\end{table}

\subsection{Sampling-Time Repair and Release}
For generator family $g$, the repair function $\mathcal{A}_g$ produces two alternatives from the original row:
\begin{equation}
\mathcal{C}_g(x_0)=\{x_0\}\cup \mathcal{A}_g(x_0).
\label{eq:candidates}
\end{equation}
The default candidate set therefore contains three records. PEG-Tab selects from this finite set using
\begin{equation}
\pi_\lambda(x\mid\mathcal{C}_g)=
\frac{\exp[-\lambda E(x)]}
{\sum_{x'\in\mathcal{C}_g}\exp[-\lambda E(x')]},
\label{eq:selection}
\end{equation}
where $\lambda$ controls the preference for lower-score records. A final threshold rejects a selected record that still exceeds the release limit. The same score and release rule are used for every generator. Only the repair mechanism changes.

PEG-Tab first scores $x_0$ and identifies the fields that contribute most to $E(x_0)$. The generator-specific repair produces two alternatives while preserving the remaining fields as much as possible. PEG-Tab then scores all three records, selects one with Equation~\ref{eq:selection}, and verifies it before release. The procedure can repeat for at most three repair rounds. Algorithm~\ref{alg:pegtab} summarises this shared procedure.

\begin{algorithm}[t]
\caption{PEG-Tab sampling-time repair and release.}
\label{alg:pegtab}
\begin{algorithmic}[1]
\REQUIRE Private table $\mathcal{D}$, fixed generator $G_\theta$, family $g$
\REQUIRE Score $E$, selection strength $\lambda$, threshold $\tau$, rounds $R=3$
\ENSURE A released record or no release
\STATE Draw $x_0 \sim G_\theta$
\FOR{$r=1,\ldots,R$}
    \STATE Identify the fields contributing most to $E(x_{r-1})$
    \STATE Generate two repairs and form $\mathcal{C}_g(x_{r-1})$
    \STATE Evaluate $E(x)$ for every $x\in\mathcal{C}_g(x_{r-1})$
    \STATE Draw $\hat{x}\sim\pi_\lambda(\cdot\mid\mathcal{C}_g(x_{r-1}))$
    \IF{$E(\hat{x})\leq\tau$}
        \RETURN $\hat{x}$
    \ENDIF
    \STATE $x_r\leftarrow\hat{x}$
\ENDFOR
\RETURN no release
\end{algorithmic}
\end{algorithm}

\paragraph{Example.}
Consider a generated row with a rare occupation and age combination that also lies close to a training record in its continuous fields. The discrete and mixed-space terms both increase. The repair mechanism edits the implicated fields while anchoring the rest of the row. The release rule then compares the original row with the two repaired alternatives. This illustrates why $E(x)$ is used to compare candidate records rather than to measure population fidelity.

\paragraph{GReaT repair.}
GReaT redecodes the row with a Top $K$ logits processor. For value token $v$ at step $t$, it applies
\begin{equation}
\ell'_t(v)=\ell_t(v)-\lambda\delta_t(v),
\label{eq:token}
\end{equation}
where $\delta_t(v)$ is computed from value, substring, and combination rarity. Schema tokens are unchanged. Two completed redecodings form the repaired alternatives.

\paragraph{CTGAN and TVAE repair.}
The method marks risky fields with mask $M$ and optimises a latent code. A compact form is
\begin{align}
\min_z\quad &L_{\mathrm{cond}}(M\odot G_\theta(z),M\odot x_{\mathrm{safe}})
+\eta_1\lVert z-z_0\rVert_2^2 \nonumber\\
&+\eta_2 L_{\mathrm{pres}}((1-M)\odot G_\theta(z),(1-M)\odot x_0).
\label{eq:latent}
\end{align}
The condition loss respects feature type. The preservation term limits changes outside the risk mask. Two restart solutions form the repaired alternatives.

\paragraph{TabDDPM repair.}
TabDDPM uses masked conditional regeneration during reverse diffusion. Safe fields follow a noised anchor, risky fields move toward lower-score targets, and the remaining dimensions follow the denoiser. Two regeneration runs form the repaired alternatives. This is a tabular adaptation of masked diffusion repair \cite{lugmayr2022repaint}.

The repair procedures are generator-specific. Their common purpose is to offer alternatives to the same record-level score and release rule. The supplement specifies all score, mask, optimisation, diffusion, threshold, and stopping settings.
\section{Experimental Setup}

\paragraph{Data and splits.}
We use Adult, Default Credit, Online Shoppers, South German Credit, and Student Performance \cite{kohavi1996scaling,yeh2009comparisons,sakar2018real,groemping2019south,cortez2008student}. For each random seed, every dataset is partitioned into four mutually disjoint subsets: private records for generator training, reference records for score calibration, nonmember records for privacy auditing, and real test records used only for TSTR evaluation. Member audit records are sampled from the private training split. The reference and real test subsets do not overlap. Neither the features nor the labels of the real test records are used to calibrate the score, construct repair candidates, select hyperparameters, or make release decisions. Reference labels are not used by PEG-Tab. The same partitions are used across all compared methods. Risk-audited columns are fixed before evaluation from domain-defined quasi-identifiers and numerical fields with high cardinality. Full split sizes and audited columns appear in the supplement.

\paragraph{Generators and utility.}
The backbones are CTGAN, TVAE, TabDDPM, and GReaT with DistilGPT-2. Each method releases the same number of synthetic rows as the private training split. Utility follows the Train on Synthetic, Test on Real protocol and averages XGBoost, a linear model, and a multilayer perceptron. Classification uses AUC, while Student Performance uses $R^2$. These quantities are not directly comparable across task types. Their mean is therefore only a compact descriptive summary, with every per-dataset and per-estimator result reported in the supplement.

\paragraph{Comparators.}
Vanilla uses the unchanged sampler. Post hoc Filter generates three times the requested number of rows and rejects records above the same risk threshold. It is an outcome-oriented deployment baseline rather than a compute-matched baseline. The supplement reports wall-clock cost, accepted-row throughput, repair rounds, and rejection rates. A six-condition pipeline ablation separates repair, reranking, and final rejection. The original three-seed benchmark retains DP-Train as a training-time privacy reference. The formal reference study uses MST and AIM implemented with dpmm \cite{mckenna2021mst,mckenna2022aim,mahiou2025dpmm}. These mechanisms provide different guarantees and are not equal-guarantee competitors to PEG-Tab.

\paragraph{Predefined tuning and transfer.}
South German Credit is the only development dataset. Because the repair mechanisms act on different representations, we select one control strength per generator family. We search $\lambda\in\{0.5,1,2,5,10\}$ and select the smallest value achieving at least 50\% relative Near Copy reduction while retaining at least 98\% of Vanilla utility. If no value meets both criteria, a deterministic fallback minimises Near Copy among utility-feasible candidates. When Vanilla Near Copy is already zero, the fallback preserves zero copying and selects the highest-utility candidate. The selected values are then frozen for the four transfer datasets.

\begin{table}[t]
\centering
\small
\setlength{\tabcolsep}{4pt}
\begin{tabular}{lcl}
\toprule
Generator & $\lambda$ & Selection outcome \\
\midrule
CTGAN & 0.5 & Meets both criteria \\
GReaT & 1 & Zero-copy fallback \\
TVAE & 2 & Minimum copy under utility constraint \\
TabDDPM & 2 & Best utility with zero copy \\
\bottomrule
\end{tabular}

\caption{Guidance strengths selected on the development dataset and then frozen.}
\label{tab:lambda_rule}
\end{table}

The complete development map and all candidate values are reported in the supplementary material.

\paragraph{Audits.}
Direct targets are Exact Copy and Near Copy. Near Copy uses a normalised mixed-type distance below a threshold calibrated from training-record distances. Score-aligned diagnostics include reconstruction, Distance to Closest Record, nearest-neighbour ratio, DPI, and DOMIAS \cite{vanbreugel2023membership,ward2024dpi}. We report their maximum AUC as Worst AUC. A fixed black-box Shadow MIA is excluded from scoring, calibration, and parameter selection \cite{shokri2017membership}. The common attack pipeline covers CTGAN, TVAE, and TabDDPM, giving 12 held-out transfer cells and using eight shadow models.

\paragraph{Uncertainty and scope.}
The original core benchmark uses seeds 42, 43, and 44 with matched splits and initialisations. The expanded transfer, ablation, sensitivity, and runtime protocol fixes seed 42 across method cells to isolate control effects under the same trained generator and split. Transfer differences are paired over 16 dataset--generator configurations with 20,000 bootstrap resamples. Shadow MIA uses the 12 applicable non-GReaT configurations. Protocol-specific and per-configuration results appear in the supplement.

\section{Results}

\subsection{Main Empirical Results}

Table~\ref{tab:aggregate_results} reports the original three-seed
benchmark and is separate from the frozen transfer analysis below.
PEG-Tab reduces mean Near Copy from $0.078$ to $0.027$ and aggregate
Exact Copy to zero. These are the largest copy reductions among the
nonformal sampling controls. Worst AUC changes from $0.543$ to $0.536$,
while TPR@1\% changes from $0.011$ to $0.009$. The main aggregate effect
is therefore copy reduction rather than broad membership protection.
Post hoc Filter also removes exact copies and has the highest aggregate
Utility, but leaves Near Copy at $0.051$. The archived DP-Train row is
retained only as context for this training-time benchmark and is not
used in the frozen transfer comparison.

\begin{table*}[t]
\centering
\small
\renewcommand{\arraystretch}{1.08}
\begin{tabular*}{\textwidth}{@{\extracolsep{\fill}}lccccc@{}}
\toprule
Method
& Utility $\uparrow$
& Worst AUC $\downarrow$
& TPR@1\% $\downarrow$
& Exact Copy $\downarrow$
& Near Copy $\downarrow$ \\
\midrule

Vanilla
& $0.561 \pm 0.590$
& $0.543 \pm 0.040$
& $0.011 \pm 0.009$
& $0.005 \pm 0.019$
& $0.078 \pm 0.148$ \\

Post hoc Filter
& $\bm{0.608 \pm 0.356}$
& $0.543 \pm 0.039$
& $0.010 \pm 0.009$
& $\bm{0.000 \pm 0.000}$
& $0.051 \pm 0.145$ \\

PEG-Tab
& $0.585 \pm 0.433$
& $0.536 \pm 0.035$
& $0.009 \pm 0.008$
& $\bm{0.000 \pm 0.000}$
& $\bm{0.027 \pm 0.114}$ \\

DP-Train
& $0.023 \pm 0.668$
& $\bm{0.527 \pm 0.024}$
& $\bm{0.008 \pm 0.008}$
& $\bm{0.000 \pm 0.001}$
& $\bm{0.027 \pm 0.068}$ \\

\bottomrule
\end{tabular*}
\caption{
Aggregate results over five datasets, four generators, and three random
seeds. Each entry reports the mean and standard deviation over 60 dataset,
generator, and seed runs. The standard deviation reflects configuration and
seed variability and is not a confidence interval. Utility is a descriptive
cross-task average of classification AUC and regression $R^2$. Bold values
denote the best. Worst AUC is the maximum AUC across Reconstruction, DCR, NN-Ratio,
DPI, and DOMIAS.
}
\label{tab:aggregate_results}
\end{table*}

\subsection{Frozen Transfer and Targeted Diagnostics}

\begin{figure*}[t]
    \centering
    \includegraphics[width=0.94\textwidth]{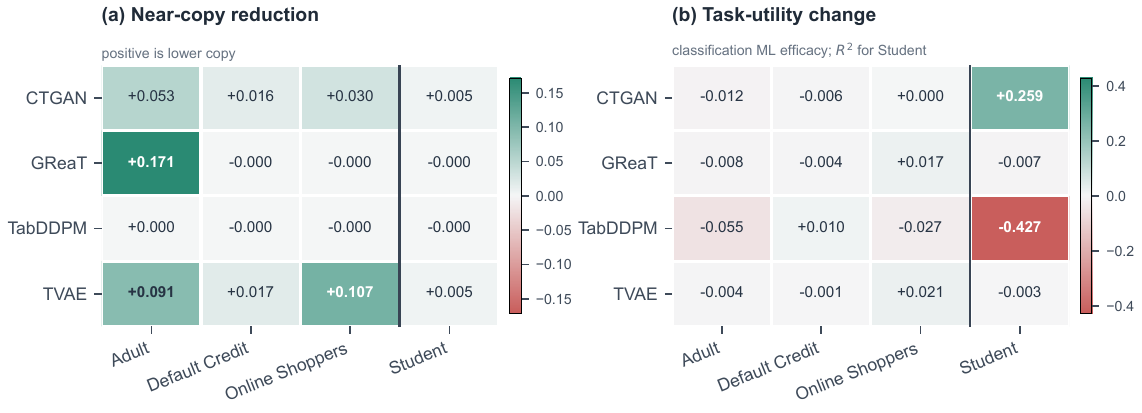}
    \caption{
    PEG-Tab transfer effects relative to Vanilla after selecting $\lambda$
    on South German Credit. Panel (a) reports the reduction in Near Copy,
    where positive values favour PEG-Tab. Panel (b) reports the change in
    task utility, where positive values favour PEG-Tab. South German Credit
    is excluded as it serves as the development dataset.
    }
    \label{fig:transfer_quilt}
\end{figure*}

The principal generalisation test selects $\lambda$ on South German
Credit and freezes it on four transfer datasets. Figure~\ref{fig:transfer_quilt}
shows that Near Copy decreases or ties in every cell, mainly through
CTGAN, GReaT, and TVAE. TabDDPM already has negligible baseline copying.
Utility is more mixed, so PEG-Tab is best applied after an audit finds
measurable copy risk.

\begin{figure*}[t]
    \centering
    \includegraphics[width=0.88\textwidth]{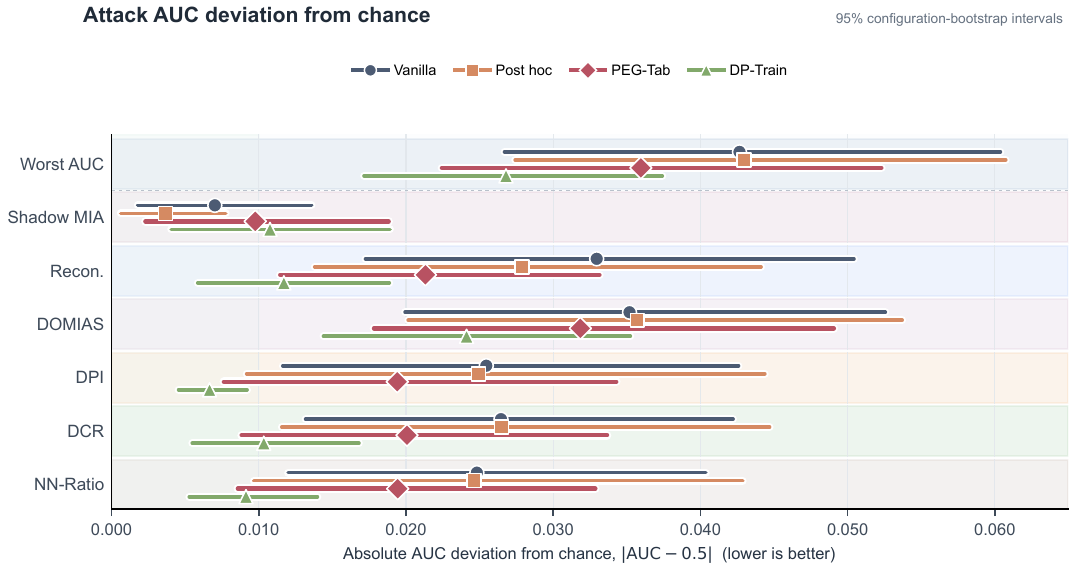}
    \caption{
    Absolute attack AUC deviation from chance. Points report the mean
    $|\mathrm{AUC}_{\mathrm{method}}-0.5|$, so values closer to zero indicate
    weaker attack discrimination. Bars show 95\% configuration-bootstrap
    intervals. Worst AUC and the five score-based attacks use 20 dataset and
    generator cells. The held-out Shadow MIA row reports the selected-lambda
    transfer protocol on 12 cells with fixed shadow-release provenance.
    }
    \label{fig:attack_deltas}
\end{figure*}

Figure~\ref{fig:attack_deltas} is a descriptive all-cell summary under
the expanded fixed-seed protocol. Relative to Vanilla, PEG-Tab reduces
the chance gap for Reconstruction, DPI, DCR, and NN-Ratio by about
$0.012$, $0.006$, $0.006$, and $0.005$. Their paired intervals remain
above zero, but these score-aligned diagnostics are corroborating rather
than independent evidence. DOMIAS decreases by $0.003$, and Shadow MIA
increases by $0.003$, with both intervals crossing zero. Worst AUC
decreases by $0.0067$. Post hoc Filter has no reliable score-aligned
reduction. 

\subsection{Pipeline Ablation and Parameter Sensitivity}

Table~\ref{tab:pipeline_ablation} isolates repair, ranking, and final
rejection on matched fixed-seed cells. No stage alone matches the full
Near Copy reduction. Repair plus ranking has the highest Utility, while
final rejection supplies the remaining copy reduction. Aligned and
Shadow AUC stay close to Vanilla, so the ablation supports stage
complementarity rather than broader attack protection.

\begin{table}[t]
\centering
\small
\setlength{\tabcolsep}{2.6pt}
\renewcommand{\arraystretch}{1.05}
\begin{tabular}{lrrrr}
\toprule
Variant & Near $\downarrow$ & Worst AUC $\downarrow$ &
Shadow $\downarrow$ & Utility $\uparrow$ \\
\midrule
Vanilla
& $0.0700$
& $\bm{0.5410}$
& $0.5059$
& $0.6116$ \\

Reject only
& $0.0664$
& $0.5432$
& $0.5082$
& $0.5566$ \\

Rerank only
& $0.0692$
& $0.5425$
& $\bm{0.5033}$
& $0.5697$ \\

Repair only
& $0.0667$
& $0.5418$
& $0.5059$
& $0.5873$ \\

Repair + rank
& $0.0609$
& $0.5440$
& $0.5065$
& $\bm{0.6145}$ \\

Full PEG-Tab
& $\bm{0.0392}$
& $0.5428$
& $0.5070$
& $0.6012$ \\
\bottomrule
\end{tabular}
\caption{
Pipeline ablation averaged over 20 dataset--generator cells.
Lower values are better except for Utility.
Worst is the maximum AUC over the five score-aligned attacks.
Bold values denote the best result across all variants.
}
\label{tab:pipeline_ablation}
\end{table}

\subsection{Comparisons with Filtering and Formal DP References}

Table~\ref{tab:filter_cells} compares PEG-Tab with Post hoc Filter.
PEG-Tab retains higher Utility in all CTGAN and TVAE transfer cells and
Pareto-dominates in five of these eight. Its advantage is smaller when
baseline copying is negligible. For TabDDPM, Near Copy ties in all four
cells and the filter has higher Utility in three. This outcome-oriented
comparison is not compute-matched, and the supplement reports the added
sampling cost.

\begin{table}[t]
\centering
\small
\setlength{\tabcolsep}{2.8pt}
\renewcommand{\arraystretch}{1.05}
\begin{tabular}{lrrr}
\toprule
Generator & Utility wins & Near B/T/W & Pareto P/F/T \\
\midrule
CTGAN
& $4/4$
& $3/0/1$
& $3/0/1$ \\

GReaT
& $3/4$
& $0/3/1$
& $2/1/1$ \\

TVAE
& $4/4$
& $1/1/2$
& $2/0/2$ \\

TabDDPM
& $1/4$
& $0/4/0$
& $1/3/0$ \\
\midrule
\textbf{All}
& $\bm{12/16}$
& $\bm{4/8/4}$
& $\bm{8/4/4}$ \\
\bottomrule
\end{tabular}
\caption{
Configuration-level comparison with Post hoc Filter using
$3\times$ oversampling on the 16 transfer cells. Utility wins count
cells where PEG-Tab has higher Utility. Near B/T/W denotes better,
tied, or worse Near Copy for PEG-Tab. Pareto P/F/T denotes PEG-Tab
dominance, filter dominance, or a trade-off.
}
\label{tab:filter_cells}
\end{table}

Table~\ref{tab:dp_refs} places PEG-Tab beside MST and AIM at two
formal privacy budgets. PEG-Tab has no formal $\varepsilon$ and is only
an empirical operating point. MST gives lower copy and attack values at
$\varepsilon=1$ with lower Utility, and the highest Utility at
$\varepsilon=8$ with higher Worst AUC. These are different guarantee
regimes rather than equal-guarantee comparisons.

\begin{table}[t]
\centering
\small
\setlength{\tabcolsep}{3.5pt}
\renewcommand{\arraystretch}{1.05}
\begin{tabular}{llrrr}
\toprule
Method
& $\varepsilon$
& Utility $\uparrow$
& Near $\downarrow$
& Worst AUC $\downarrow$ \\
\midrule
PEG-Tab
& n/a
& $0.585$
& $0.027$
& $0.536$ \\

MST
& $1$
& $0.307$
& $\bm{0.023}$
& $\bm{0.531}$ \\

MST
& $8$
& $\bm{0.630}$
& $0.026$
& $0.551$ \\

AIM
& $1$
& $0.213$
& $0.025$
& $0.546$ \\

AIM
& $8$
& $0.462$
& $0.032$
& $0.551$ \\
\bottomrule
\end{tabular}
\caption{
Contextual comparison with formal DP reference mechanisms. The
PEG-Tab row is taken from the three-seed benchmark in
Table~\ref{tab:aggregate_results}. MST and AIM are evaluated on the
same five datasets using the same utility and attack definitions.
PEG-Tab has no formal $\varepsilon$. Bold values denote the best
reported result in each column. Per-dataset results and privacy
accounting details appear in the supplement.
}
\label{tab:dp_refs}
\end{table}

\section{Discussion}

\paragraph{What the results establish.}
PEG-Tab provides a post-training control point for copied records and
close training neighbours. The clearest evidence is the frozen transfer
analysis, where Near Copy decreases or ties in every cell after selecting
guidance on one development dataset. Exact Copy provides supporting
evidence, although most transfer configurations already contain no exact
copies. Reconstruction, DPI, DCR, and NN-Ratio also move favourably under
the chance-adjusted analysis. These diagnostics are closely aligned with
the record score, so their improvements are corroborating rather than
independent evidence. DOMIAS and the held-out Shadow MIA show no minor
change. This boundary is important. The results support targeted control
of copy-style memorisation and do not establish a general membership
inference defence.

\paragraph{Why the shared release rule matters.}
Each backbone repairs a row differently. GReaT edits the decoding path,
CTGAN and TVAE edit latent representations, and TabDDPM edits selected
fields during reverse sampling. After these model-specific repairs, every
candidate is evaluated by the same calibrated score and release rule. The
ablation supports this division of labour. Repair creates alternatives,
ranking chooses among them, and final rejection removes residual failures.
No stage alone explains the full Near Copy reduction. This separation is
the main reusable element of PEG-Tab because a new generator only requires
a native repair operator while retaining the common release interface.

\paragraph{When sampling-time repair is useful.}
PEG-Tab complements rather than uniformly replaces inexpensive filtering.
Compared with Post hoc Filter, it retains higher utility in 12 of 16
transfer cells and Pareto-dominates the filter in eight, while the filter
dominates in four. The benefit is strongest when copying is measurable and
rejection discards otherwise useful records. When the base generator has
negligible Near Copy, as in several TabDDPM settings, leaving the sampler
unchanged or using a simple filter is preferable. The comparison is
outcome-oriented rather than compute-matched because repair requires
additional model-specific inference. Runtime, release yield, and the
required copy-risk operating point should therefore be audited together.

\paragraph{Interpreting the score.}
The record score is a release decision aid, not a new privacy definition.
It checks whether a candidate copies rare content, lies close to a training
row, or is unusually concentrated around training data. Improvements on
related diagnostics are expected to reflect this construction. A held-out
attack is needed to identify the boundary of the result, and broader
evaluations should include attacks that are not used for scoring or tuning.
Rarity is also a potentially consequential signal. Deployment should audit
whether repair disproportionately changes rare categories, tail records,
or small subgroups even when aggregate task utility remains stable.

\paragraph{Limitations.}
PEG-Tab requires the private training table, a permissible reference
population, and generator-specific inference states. The original benchmark
uses three seeds, while the expanded transfer, ablation, sensitivity, and
runtime protocol uses one matched seed to isolate controlled effects. The
Shadow analysis covers 12 non-GReaT configurations and does not test
repeated releases or adaptive queries. The study therefore does not
establish protection against broader membership, attribute inference,
linkage, or auxiliary-information attacks. Sampling cost may also limit
interactive use, particularly for latent optimisation and reverse diffusion.

\section{Conclusion}

PEG-Tab adds a sampling-time repair and release-control layer to frozen tabular generators. Generator-native edits propose alternatives, while one calibrated score and gate control release across GReaT, CTGAN, TVAE, and TabDDPM. The clearest result is lower Exact Copy and Near Copy, with repair often preserving more utility than filtering when copying is measurable. Unchanged DOMIAS and held-out Shadow MIA limit the contribution to copy-style memorisation rather than formal or general privacy protection.

\clearpage
\IfFileExists{aaai2027.bst}{\bibliographystyle{aaai2027}}{\bibliographystyle{aaai-named}}
\bibliography{references}
\end{document}